\documentclass[runningheads]{llncs}
\usepackage[T1]{fontenc}
\usepackage{graphicx,verbatim}
\usepackage{amsmath}
 \usepackage{booktabs,multirow,subcaption}
 \usepackage{hyperref}
 \usepackage{orcidlink}
 \hypersetup{
    colorlinks,
    linkcolor={red!50!black},
    citecolor={blue!50!black},
    urlcolor={blue!50!black}
}
\usepackage{color}

\begin{document}
\title{Wasserstein Equilibrium Decoding for Reliable Medical Visual Question Answering}
\titlerunning{Wasserstein Equilibrium Decoding}
%

\author{Luca Hagen\inst{1}\orcidlink{0009-0005-0990-6807} \and
Johanna P. Müller\inst{1}\orcidlink{0000-0001-8636-7986} \and
Weitong Zhang\inst{2}\orcidlink{0000-0002-3681-4546} \and
Mengyun Qiao\inst{3}\orcidlink{0000-0002-5157-1079} \and
Bernhard Kainz\inst{1,2}\orcidlink{0000-0002-7813-5023}}
\authorrunning{L. Hagen et al.}
\institute{
Friedrich-Alexander University Erlangen-Nürnberg, Erlangen, Germany \and
Imperial College London, London, United Kingdom \and
University College London, London, United Kingdom\\[0.5em]
Corresponding author: \texttt{luca.hagen@fau.de}
}

\maketitle              
\begin{abstract}

Small vision-language models (2-8B) are well-suited for clinical deployment due to privacy constraints, limited connectivity, and low-latency requirements favouring on-device or on-premise inference. However, their limited capacity exacerbates the generation of plausible but incorrect outputs.
We extend game-theoretic decoding, previously restricted to text-only, closed-ended NLP tasks, to vision-language models for open-ended Medical VQA. We introduce a semantically aware Wasserstein stopping criterion that replaces lexical order matching, enabling convergence based on semantic consensus among near-synonymous candidate answers and avoiding unnecessary iterations caused by clinically equivalent ranking swaps. On VQA-RAD and PathVQA, we obtain consistent, statistically significant improvements over greedy and discriminative baselines. On VQA-RAD, we improve Qwen3-VL-2B by $+3.5$ percentage points (p < 0.01), surpassing the greedy 4B model, with similar trends at larger scales. On PathVQA, Gemma-3-4B with BDG matches MedGemma-4B under greedy decoding despite no domain-specific fine-tuning. At accuracy parity with classic BDG, the Wasserstein criterion reduces average convergence iterations by approximately 20\%, improving inference efficiency while preserving the game-theoretic equilibrium behaviour. Code is available at \url{https://github.com/luca-hagen/Wasserstein-BDG-medical-VQA}.

\keywords{VQA  \and Vision-Language Models \and Game Theory.}

\end{abstract}
\section{Introduction}

Medical Visual Question Answering (Med-VQA) requires models to jointly reason over medical images and natural-language queries to produce concise, correct answers, with applications in radiology report triage, pathology grading, and clinical decision support~\cite{lau2018dataset,he2020pathvqa,lin2023medvqasurvey}.
Recent vision-language models (VLMs) have advanced the state of the art on standard Med-VQA benchmarks~\cite{li2024llava-med,moor2023medflamingo}, yet most high-performing systems require cloud-hosted GPU infrastructure for inference.
In clinical practice this is frequently infeasible: data protection regulations require that patient data remain on premises~\cite {murdoch2021privacy}, rural and point-of-care settings may lack connectivity for real-time cloud access, and latency-sensitive applications demand immediate responses.
These constraints are driving a shift toward small VLMs (1-8B parameters) deployable on local hardware~\cite{abdin2024phi3,garg2026rise}, making their reliability for medical reasoning a practical requirement for the next generation of on-device clinical AI.

Small VLMs, however, inherit and amplify the tendency of language models to produce plausible but factually incorrect outputs~\cite{zhu-etal-2025-trust,zhang2025siren}.
In the medical domain, where a confidently wrong answer can directly influence patient safety, this is particularly problematic in \emph{open-ended} settings where models must generate free-form responses without the guidance of predefined options, such as real clinical queries.
A natural direction for improving reliability without retraining is intervention at inference time, but existing approaches each have known failure modes: self-reflection suffers from confirmation bias~\cite{huang2024largelanguagemodelsselfcorrect}, self-consistency~\cite{wang2023selfconsistencyimproveschainthought} provides no discriminative verification, and multi-agent debate is prone to reinforced error propagation~\cite{du2024improving}.

Game-theoretic decoding offers a principled alternative.
\cite{jacob2023consensus} proposed the \emph{Equilibrium Consensus Game} (ECG), framing inference as a signalling game between a generator and verifier that converge to a Nash equilibrium~\cite{nash1950equilibrium}.
\cite{zhang2025from} identified \emph{collusion} as a key weakness of ECG and proposed the \emph{Bayesian Decoding Game} (BDG), which enforces a $\sigma$-separated equilibrium condition to avoid it, enabling smaller models to outperform models an order of magnitude larger through the game mechanism alone.
Both frameworks, however, were designed for text-only models on closed-ended tasks. Thus, our goal is not to replace BDG, but to extend it to a regime in which its original assumptions no longer hold: multimodal inputs, open-ended answer spaces, and candidate sets containing clinically near-equivalent phrasings.
In this setting, exact lexical rank agreement can block convergence even when generator and verifier agree semantically, because the model may split probability mass across synonymous candidates such as \emph{liver} and \emph{hepatic region}.
Enabling BDG in open-ended Med-VQA with small VLMs therefore requires addressing three challenges: (1)~generator and verifier must operate over joint vision-language inputs; (2)~the finite candidate set must be constructed dynamically; and (3)~open-ended sampling yields semantically near-duplicate candidates (\emph{e.g.}, \emph{liver} vs.\ \emph{hepatic region}), causing the lexical order-match stopping criterion to penalise semantically irrelevant ranking swaps. Our contributions are:
\begin{enumerate}
    \item We extend BDG to vision-language inputs, reformulating generator and verifier for multimodal Medical VQA.
    \item We propose temperature-diverse sampling with semantic deduplication to construct a finite candidate set, enabling BDG in open-ended settings.
    \item We introduce a semantically-aware stopping criterion based on Wasserstein-1 distance~\cite{rubner2000earth}, allowing convergence when generator and verifier agree up to clinically near-equivalent candidate phrasings rather than exact lexical rank order.
    \item We demonstrate that BDG substantially improves small general-purpose VLMs on VQA-RAD~\cite{lau2018dataset} and PathVQA~\cite{he2020pathvqa}, in some settings matching or exceeding larger or domain-specialised baselines, while the Wasserstein criterion reduces convergence iterations by approximately 20\% compared to classic BDG.
\end{enumerate}

\noindent\textbf{Related Work.}  
\emph{Medical VQA} has evolved from discriminative methods~\cite{lin2023medvqasurvey} to generative VLM-based approaches~\cite{dong2025generative,hartsock2024visionlanguage} across datasets like VQA-RAD~\cite{lau2018dataset}, PathVQA~\cite{he2020pathvqa}, and PMC-VQA~\cite{zhang2024pmcvqa}. Domain-specialised VLMs such as LLaVA-Med~\cite{li2024llava-med} and Med-Flamingo~\cite{moor2023medflamingo} achieve strong results via biomedical fine-tuning, while general-purpose models like Qwen2-VL~\cite{qwen3technicalreport} and Phi-3-Vision~\cite{abdin2024phi3} can be competitive through high-quality training and instruction tuning.  
\emph{Hallucination} in LLMs and VLMs~\cite{zhang2025siren,zhu-etal-2025-trust,ji2023hallucinationsurvey} poses patient safety risks. Inference-time correction strategies include self-consistency~\cite{wang2023selfconsistencyimproveschainthought}, contrastive decoding~\cite{li2023contrastive}, internal consistency checks~\cite{liang2024internalconsistencyselffeedbacklarge}, and multi-agent debate~\cite{du2024improving,wang2024rethinking}, each with limitations such as confirmation bias.  
\emph{Game-theoretic decoding}~\cite{jacob2023consensus,zhang2025from} with collusion-aware constraints~\cite{bonjour2022information,xu2024mechanism,koirala2024algorithmic} has so far been applied only to text-only, closed-ended models. \emph{Small language models for healthcare}~\cite{garg2026rise,abdin2024phi3,team2024gemma} are motivated by privacy and low-latency needs~\cite{chen2025toward,murdoch2021privacy}. Semantic similarity using biomedical encoders such as SapBERT~\cite{liu2021sapbert} enables fine-grained answer evaluation. To our knowledge, we are the first to extend game-theoretic decoding to multimodal VLMs, open-ended generation, and small models suitable for clinical edge deployment.

\begin{figure}[h]
    \centering
    \includegraphics[width=0.99\textwidth]{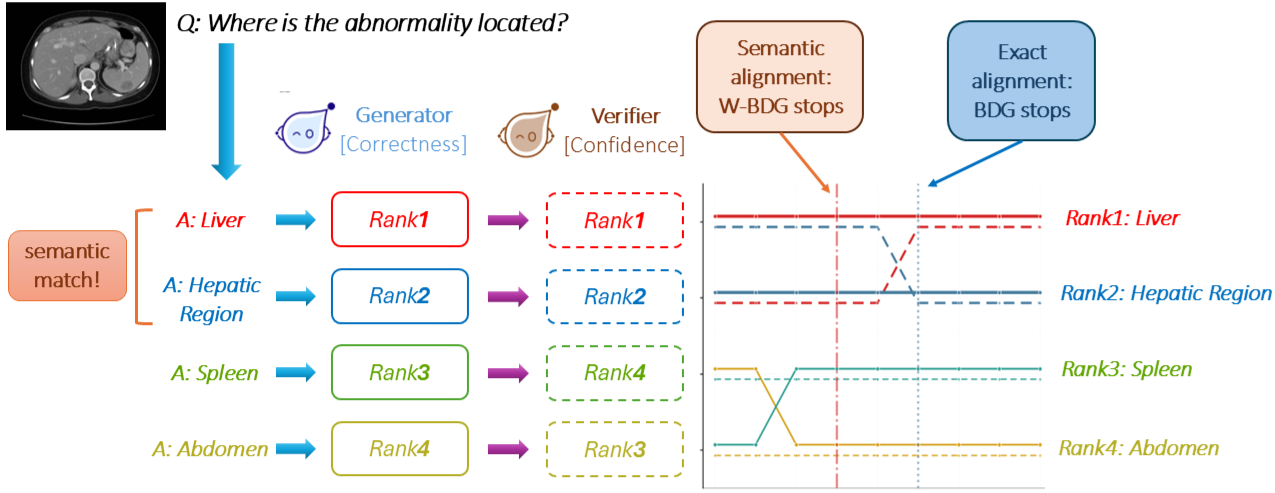}
    \caption{\textbf{Wasserstein-BDG for open-ended medical VQA.} Given an image and question, the Generator produces candidate answers, which the Generator (solid) and Verifier (dashed) iteratively align via game-theoretic updates. W-BDG converges at semantic consensus, allowing swaps between near-synonymous answers (e.g., Liver $\cong$ Hepatic Region), whereas classic BDG requires exact rank agreement.}
    \label{fig:scheme}
\end{figure}

\section{Method}
\label{sec:bdg}
We build on the BDG~\cite{zhang2025from} and the game-theoretic formulation introduced by \cite{jacob2023consensus},
which formalises LLM decoding as a two-player signalling game.

\noindent\textbf{Signaling game.}
A signalling game~\cite{gibbons1992primer} consists of a sender who observes a private environment signal and takes an action to transmit information to a receiver, who must infer the signal from the action alone. 
In the BDG, this game is defined as: given a query $x$ and a finite candidate set $Y = \{y_1, \ldots, y_n\}$, the \emph{generator} $G$ receives a signal $s \in \{\textit{correct}, \textit{incorrect}\}$ and selects a action distribution over $Y$, while the \emph{verifier} $V$ observes the generator's action and judges each candidate as correct or incorrect. Both agents operate under partial information: the generator does not know the verifier's internal scores, while the verifier does not observe the original signal.
Formally, the generator's strategy is $a_G(\cdot \mid x, s) \in \Delta(Y)$ and the verifier's strategy is $a_V(\cdot \mid x, y) \in \Delta(\{\textit{correct}, \textit{incorrect}\})$.
The shared utility is determined by preference ordering agreement:
\begin{equation}
    u_G(\pi_G, \pi_V) \;=\; u_V(\pi_G, \pi_V) \;=\;
    \frac{1}{2} \sum_{s \,\in\, S} \mathbf{1}\bigl[O_G = O_V \mid s\bigr],
\end{equation}
where $O_G, O_V \in \mathcal{P}_{|Y|}$ are the preference orderings induced by the respective action distributions. To prevent \emph{collusion}, an undesirable equilibrium in which both agents mutually reinforce an incorrect answer rather than correcting it, BDG constrains the solution to a \emph{$\sigma$-Separated Equilibrium} ($\sigma$-SE). This requires the generator to maintain a minimum separation of $\sigma_G$ between its correct and incorrect distributions, and the verifier to maintain a
separation of $\sigma_V$ between its correctness scores for each candidate:
$
    \min_{y \,\in\, Y}\;
        \bigl\| a_G(y \mid x,\, \textit{correct}) - a_G(y \mid x,\, \textit{incorrect}) \bigr\|
        > \sigma_G; 
    \min_{y \,\in\, Y}\;
        \bigl\| a_V(\textit{correct} \mid x, y) - a_V(\textit{incorrect} \mid x, y) \bigr\|
        > \sigma_V.
$
\noindent\textbf{Markovian strategy update.}
Starting from initial policies $a_G^{(1)}$ and $a_V^{(1)}$ derived from the models token probabilities, both agents update their strategies through a Markovian no-regret schedule. At each iteration $t$, each agent adapts based on the opponent's previous action:
\begin{align}
    a_G^{(t+1)}(y \mid x, s) \;&\propto\; \exp\!\left(
        \frac{
            \tfrac{1}{2}\,\tilde{a}_V^{(t)}(s \mid x, y)
            \;+\; \lambda_G \log a_G^{(t)}(y \mid x, s)
        }{
            1/(\eta_G\, t) + \lambda_G
        }
    \right), \label{eq:update_G} \\[6pt]
    a_V^{(t+1)}(s \mid x, y) \;&\propto\; \exp\!\left(
        \frac{
            \tfrac{1}{2}\,\tilde{a}_G^{(t)}(y \mid x, s)
            \;+\; \lambda_V \log a_V^{(t)}(s \mid x, y)
        }{
            1/(\eta_V\, t) + \lambda_V
        }
    \right), \label{eq:update_V}
\end{align}
where $\tilde{a}_V^{(t)}$ and $\tilde{a}_G^{(t)}$ are normalized distribtuions of the opponent's current action, $\eta_G, \eta_V > 0$ are learning rates, and $\lambda_G, \lambda_V > 0$ are stiffness parameters controlling the influence of the prior policy relative to the opponent's signal. This update schedule converges to an optimal $\sigma$-SE under mild rationality conditions~\cite{zhang2025from}.

\noindent\textbf{Initialisation and original stopping criterion.}
The generator is initialised by teacher-forced scoring of each candidate $y_i \in Y$ under two prompt conditions corresponding to the signals $s \in \{\textit{correct},\textit{incorrect}\}$.
In canonical form, the correct prompt is
\emph{``Question: \{Q\} Answer: \{$y_i$\} Is this answer correct? Response: correct''},
with the incorrect prompt identical except for the final token \emph{incorrect}; the full templates include domain-specific framing and are provided in the anonymised repository.
The teacher-forced log-probability of the final label token yields
$\log P_\mathrm{LM}(y_i \mid x, s)$, and softmax normalisation over candidates gives the initial generator strategy $a_G^{(1)}(\cdot \mid x,s)$.
The verifier reverses the conditioning: given $(x,y_i)$, it scores the label tokens \textit{correct} and \textit{incorrect}, yielding $a_V^{(1)}(\cdot \mid x,y_i)$ for each candidate.
BDG declares convergence once the $\sigma$-separation condition is satisfied and the preference orderings of generator and verifier match, $O_G = O_V$.  
This exact order-match criterion is ill-suited for open-ended settings where the candidate set naturally contains semantically similar candidates.  
Since BDG operates over a finite candidate set $Y$, applying it to open-ended VQA requires dynamically constructing $Y$ from the model's output distribution. We approximate this set via sampling, followed by semantic deduplication.

\noindent\textbf{Temperature-diverse sampling.}  
Given a multimodal query $x=(I,Q)$ with image $I$ and a textual question $Q$, we sample up to 8 candidate answers from the VLM using nucleus sampling at temperatures $\mathcal{T} = \{0.5, 1.0\}$. Each sample is canonicalised by removing filler prefixes, punctuation, and extra sentences, keeping only the core answer. Sampling continues until 8 unique candidates are obtained or the maximum calls are reached, forming $Y = \{y_1, \ldots, y_n\}$. Varying temperatures capture both high-confidence and lower-probability answers, naturally including semantically similar candidates (e.g., \emph{liver} and \emph{hepatic region}) that reflect the model’s belief distribution.

\noindent\textbf{Ground metric.}
We embed all candidates $y_i \in Y$ using a biomedical concept encoder, producing embeddings that place similar medical concepts close in space, and compute a pairwise semantic distance matrix $D$ via cosine similarity:  

$D$ defines the ground metric for the Wasserstein stopping criterion and semantic evaluation, assigning small distances to near-synonymous candidates (e.g., \emph{liver} and \emph{hepatic region}) and large distances to distinct ones (e.g., \emph{liver} and \emph{spleen}).


\noindent\textbf{Semantic convergence via Wasserstein-1 distance.}  
The original BDG stopping criterion ($O_G = O_V$) requires exact agreement between the generator and verifier rankings, which is appropriate for closed-ended tasks with semantically distinct candidates. In open-ended settings, however, near-synonymous candidates are common. Lexical order-match treats swaps between semantically similar answers (e.g., \emph{liver} and \emph{hepatic region}) the same as swaps between distinct ones (e.g., \emph{liver} and \emph{spleen}), making convergence dependent on candidate form rather than meaning. This is particularly problematic because open-ended sampling can fragment the model's evidence across several clinically equivalent surface forms.
In such cases, the game may have already concentrated mass on the correct semantic concept, while exact rank-match still fails due to arbitrary ordering among synonyms.  
To address this, we replace order-match with a convergence measure based on the Wasserstein-1 distance~\cite{rubner2000earth} under the semantic ground metric $D$. Given the generator's distribution $p_G^{(t)} = a_G^{(t)}(\cdot \mid x, \textit{correct})$ and the verifier's normalized scores $p_V^{(t)} = a_V^{(t)}(\textit{correct} \mid x, y_i) / \sum_j a_V^{(t)}(\textit{correct} \mid x, y_j)$, the Wasserstein-1 distance is  
\begin{equation}
    W_1\bigl(p_G^{(t)},\, p_V^{(t)},\, D\bigr)
    = \min_{\gamma \in \Pi(p_G^{(t)},\, p_V^{(t)})} \sum_{i,j} \gamma_{ij}\, D_{ij},
    \label{eq:w1}
\end{equation}
where $\gamma_{ij}$ denotes the mass transported from candidate $y_i$ to $y_j$, and $\Pi(p_G^{(t)}, p_V^{(t)})$ is the set of valid transport plans with marginals $p_G^{(t)}$ and $p_V^{(t)}$. Intuitively, $W_1$ measures the optimal transport from generator to verifier, assigning low cost to moving mass between semantically similar candidates and high cost to semantically distant ones.

\noindent\textbf{Separation-weighted criterion.}  
To prevent premature termination of the Wasserstein convergence, we weight $W_1$ by the current $\sigma$-separation of the agents:
\begin{equation}
    \tilde{W}_1^{(t)} = \frac{W_1(p_G^{(t)},\, p_V^{(t)},\, D)}{\sigma_V^{(t)} + \varepsilon},
    \label{eq:w1_sep}
\end{equation}
where $\sigma_V^{(t)} = \min_{y} \|a_V^{(t)}(y \mid x, \textit{correct}) - a_V^{(t)}(y \mid x, \textit{incorrect})\|$ and $\varepsilon>0$ ensures numerical stability. Early in the game, small separation keeps $\tilde{W}_1^{(t)}$ large, preventing premature termination. Once distributions are well-separated and semantically aligned, $\tilde{W}_1^{(t)}$ falls below the threshold $\delta_W$, allowing termination when $\tilde{W}_1^{(t)} < \delta_W$. This relaxes exact order-match to \emph{semantic consensus}: remaining ranking disagreements are tolerated only when they involve semantically similar candidates under $D$ and concern little probability mass, i.e.\ each unresolved swap contributes a small transport cost $\gamma_{ij} D_{ij}$. High-mass disagreements between distant candidates must still be resolved before termination.

\section{Experiments}

\begin{figure}[th!]
\centering
\includegraphics[height=0.175\textheight,width=0.99\textwidth]{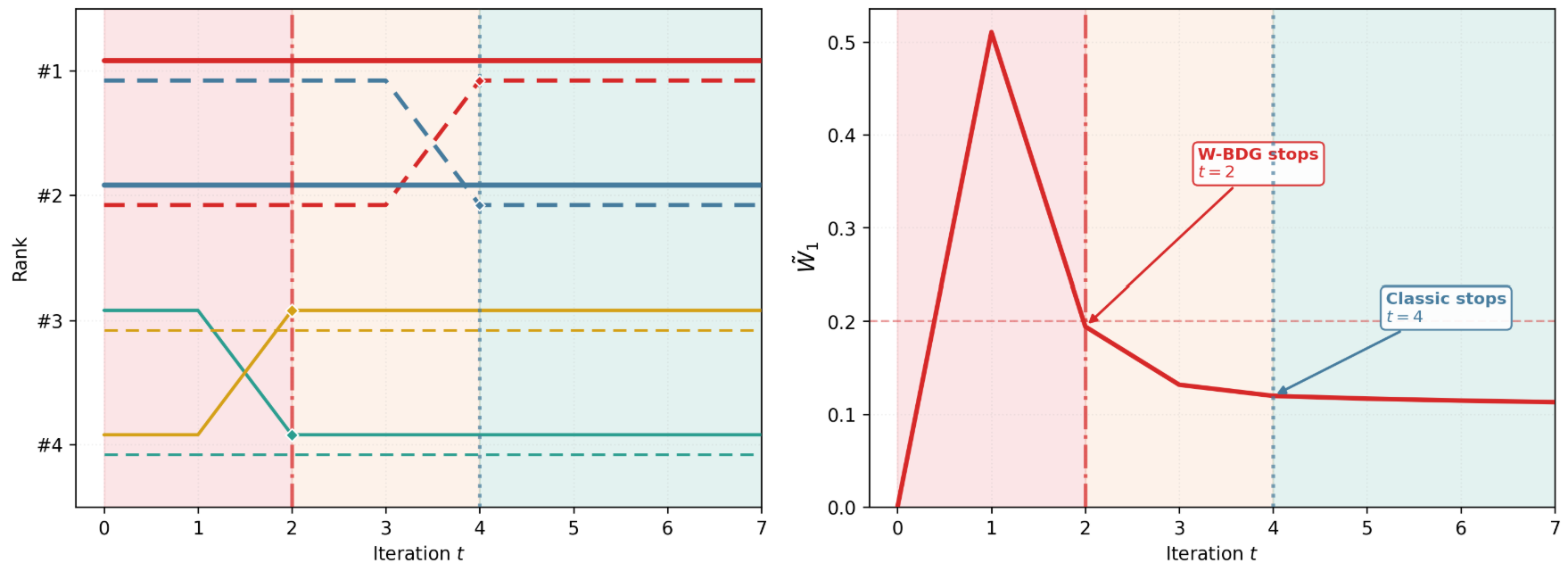}
\caption{\textbf{Convergence analysis on VQA-RAD (Qwen3-VL-4B).} \textbf{(a)}~Preference rankings over game iterations. In the red phase, rankings diverge; in the orange phase, only semantically close candidates remain swapped; in the green phase, exact rank agreement is reached. \textbf{(b)}~Separation-weighted Wasserstein distance $\tilde{W}_1^{(t)}$ over iterations. BDG-W terminates once $\tilde{W}_1^{(t)} < \delta_W$ (dashed horizontal line), tolerating the remaining semantic swap ($t{=}3$), while BDG-Classic requires exact order-match ($t{=}4$).}
\label{fig:convergence}
\end{figure}

\noindent\textbf{Datasets.}  
We evaluate on two open-ended Medical VQA benchmarks. \textbf{VQA-RAD}~\cite{lau2018dataset} has 315 radiology images and 3,515 clinician-generated QA pairs covering modality, anatomy, and abnormalities. \textbf{PathVQA}~\cite{he2020pathvqa} contains 4,998 pathology images with 32,799 QA pairs. Evaluation uses the open-ended subset of each test split, excluding yes-no questions, which reduce to binary classification and are irrelevant for BDG.

\noindent\textbf{Models and implementation.}  
We evaluate general-purpose instruction-tuned VLMs: \textbf{Gemma-3-4B}~\cite{team2024gemma}, \textbf{Qwen3-VL-2B}, \textbf{Qwen3-VL-4B}, and \textbf{Qwen3-VL-8B}~\cite{qwen3technicalreport}, alongside the domain-specialised \textbf{MedGemma-4B} (Gemma-3-4B fine-tuned on medical image-text data). Greedy decoding is used for generator baselines, while BDG candidates are generated via temperature-diverse sampling. SapBERT~\cite{liu2021sapbert} provides embeddings, and generator/verifier are initialised from token-level log-probabilities over $Y$~\cite{zhang2025from}.  
Candidates are generated with $N_\text{max}=16$ and $n=8$. BDG hyperparameters are $\lambda_G=\lambda_V=0.4$, $\eta_G=\eta_V=0.4$, $\sigma=0.005$, $\delta_W=0.2$, with a maximum of $T=500$ iterations. All experiments are run on a single NVIDIA A100 80GB GPU.

\noindent\textbf{Baselines and Evaluation metrics.}  
We compare five decoding strategies for each model:  
(i) \textbf{Greedy} (G), standard single-pass argmax;  
(ii) \textbf{Self-Contrastive Decoding} (SCD)~\cite{jacob2023consensus}, which contrasts likelihoods under correct vs incorrect prompts;  
(iii) \textbf{Verifier-only} (V), selecting the candidate with the highest initial verifier score $a_V^{(1)}(\textit{correct} \mid x, y_i)$;  
(iv) \textbf{BDG-Classic}, using the original lexical order-match stopping criterion ($O_G = O_V$);  
(v) \textbf{BDG-W} (ours), using the semantically aware $\tilde{W}_1$ criterion. Both BDG variants share the same candidate set, initialisation, and update schedule, differing only in the stopping criterion.  
We evaluate using \textbf{exact match} accuracy, \textbf{token-level F1}, and \textbf{semantic similarity} (cosine similarity between SapBERT embeddings of predicted and ground-truth answers). We also report \textbf{judge accuracy}, assessed by a VLM-as-judge (Grok 4.1 Fast~\cite{xai2025grok41}) that explicitly accounts for synonyms and clinically equivalent phrasings.

\noindent\textbf{Results.} 
Table~\ref{tab:main} reports performance across all models and decoding strategies on VQA-RAD and PathVQA.
BDG-W yields improvements across nearly all model-dataset combinations, with diminishing returns at 8B scale on VQA-RAD (Table~1).
On VQA-RAD, BDG-W improves judge accuracy of Qwen3-VL-2B by $3.5$ percentage points over greedy decoding ($p < 0.01$, paired bootstrap test~\cite{koehn2004statistical}) surpassing the greedy decoding of Qwen3-VL-4B, a model double its size. Similar behaviour can be observed from Qwen-VL-4B to Qwen-VL-8B.
Notably, Gemma-3-4B with BDG (both Wasserstein and classic) exceeds MedGemma-4b with greedy decoding on PathVQA dataset despite no domain-specific fine-tuning.
The performance of BDG-W and BDG-Classic is consistent across all settings, indicating that the Wasserstein criterion preserves the accuracy of classic BDG while avoiding unnecessary iterations caused by semantically negligible ranking disagreements.

\begin{table}[th!]
\begin{minipage}[t]{0.47\textwidth}
\centering
\caption{\textbf{Relative gain ($\Delta$\,Judge) of BDG-W over Greedy by model scale.} Gains diminish with size, indicating the game is most effective for smaller models.}
\label{tab:scaling}
\vspace{2pt}
\begin{tabular}{@{}lcc@{}}
\toprule
\textbf{Model} & \multicolumn{2}{c}{$\Delta$\,\textbf{Judge}} \\
\cmidrule(l){2-3}
(Params) & RAD & Path \\
\midrule
Qwen-2B  & $+3.5$ & $+2.2$ \\
Qwen-4B  & $+3.8$ & $+3.2$ \\
Qwen-8B  & $-1.2$ & $+1.2$ \\
Gemma-4B  & $+2.3$ & $+5.0$ \\
MedGemma-4B  & $+2.7$ & $+2.0$ \\
\bottomrule
\end{tabular}
\end{minipage}%
\hfill
\begin{minipage}[t]{0.50\textwidth}
\centering
\caption{\textbf{Convergence.} Number of iterations until convergence on VQA-RAD by W-BDG and Classic BDG. Values are mean$_{\pm\text{std}}$ over 5 runs with different random seeds.}
\label{tab:ablation}
\vspace{2pt}
\vspace{2pt}
\begin{tabular}{@{}lcc@{}}
\toprule
\textbf{Model} & \multicolumn{2}{c}{\textbf{\#Iterations}} \\
\cmidrule(l){2-3}
& W-BDG & Classic BDG \\
\midrule
Qwen-2B  & $27.46_{\pm 01.52}$ & $32.68_{\pm 00.95}$ \\
Qwen-4B  & $24.14_{\pm 01.49}$ & $27.80_{\pm 01.16}$ \\
Qwen-8B  & $19.10_{\pm 00.93}$ & $22.84_{\pm 00.99}$ \\
Gemma-4B & $16.12_{\pm 01.68}$ & $22.18_{\pm 00.53}$ \\
MedGemma-4B & $14.02_{\pm 01.95}$ & $20.62_{\pm 01.09}$ \\
\bottomrule
\end{tabular}
\end{minipage}
\end{table}

\begin{table}[!htbp]
\centering
\caption{\textbf{Main results on VQA-RAD and PathVQA (open-ended subsets).} We report exact match accuracy (EM), token-level F1, semantic similarity (Sem), and judge accuracy (Judge). Values are mean$_{\pm\text{std}}$ over 5 runs with different random seeds. Best per model in \textbf{bold}; best overall \underline{underlined}. $^\dagger$ denotes a statistically significant improvement over the corresponding Greedy baseline for the same model and dataset ($p{<}0.01$, paired bootstrap, 10k resamples).}
\label{tab:main}
\setlength{\tabcolsep}{3.2pt}
\renewcommand{\arraystretch}{1.08}
\resizebox{\textwidth}{!}{%
\begin{tabular}{@{}ll cccc cccc@{}}
\toprule
& & \multicolumn{4}{c}{\textbf{VQA-RAD}} & \multicolumn{4}{c}{\textbf{PathVQA}} \\
\cmidrule(lr){3-6} \cmidrule(lr){7-10}
\textbf{Model} & \textbf{Method} & EM & F1 & Sem & Judge & EM & F1 & Sem & Judge \\
\midrule
\multirow{5}{*}{\shortstack[l]{Qwen3-VL\\2B}}
 & Greedy       & $17.50_{\pm 00.00}$ & $26.73_{\pm 00.00}$ & $59.36_{\pm 00.00}$ & $33.50_{\pm 00.00}$ & $01.49_{\pm 00.00}$ & $03.79_{\pm 00.00}$ & $36.32_{\pm 00.00}$ & $04.98_{\pm 00.00}$ \\
 & SCD          & $15.35_{\pm 00.74}$ & $23.33_{\pm 00.94}$ & $58.62_{\pm 00.48}$ & $30.79_{\pm 00.86}$ & $00.80_{\pm 00.24}$ & $03.02_{\pm 00.46}$ & $35.38_{\pm 00.36}$ & $02.79_{\pm 00.67}$ \\
 & Verifier     & $17.25_{\pm 01.10}$ & $27.71_{\pm 00.83}$ & $\mathbf{60.82_{\pm 00.85}}$ & $35.71_{\pm 00.83}$ & $\mathbf{02.59_{\pm 00.73}}$ & $\mathbf{05.71_{\pm 00.59}}$ & $\mathbf{37.37_{\pm 00.39}}$ & $06.57_{\pm 01.23}$ \\
 & BDG-Classic  & $19.06_{\pm 00.75}$ & $28.58_{\pm 00.44}$ & $60.60_{\pm 00.54}$ & $36.51_{\pm 00.84}$ & $02.29_{\pm 00.60}$ & $04.77_{\pm 00.53}$ & $36.88_{\pm 00.26}$ & $06.93_{\pm 00.37}$ \\
 & \textbf{BDG-W} & $\mathbf{19.66_{\pm 00.63}}^{\dagger}$ & $\mathbf{28.83_{\pm 00.40}}^{\dagger}$ & $60.61_{\pm 00.64}^{\dagger}$ & $\mathbf{37.04_{\pm 00.94}}^{\dagger}$ & $02.49_{\pm 00.44}^{\dagger}$ & $05.05_{\pm 00.50}^{\dagger}$ & $36.98_{\pm 00.30}^{\dagger}$ & $\mathbf{07.17_{\pm 00.67}}^{\dagger}$ \\
\midrule
\multirow{5}{*}{\shortstack[l]{Qwen3-VL\\4B}}
 & Greedy       & $15.50_{\pm 00.00}$ & $25.63_{\pm 00.00}$ & $59.74_{\pm 00.00}$ & $36.50_{\pm 00.00}$ & $02.49_{\pm 00.00}$ & $06.65_{\pm 00.00}$ & $37.59_{\pm 00.00}$ & $07.96_{\pm 00.00}$ \\
 & SCD          & $16.40_{\pm 00.58}$ & $24.98_{\pm 00.61}$ & $59.92_{\pm 00.67}$ & $36.60_{\pm 01.39}$ & $02.09_{\pm 00.37}$ & $05.89_{\pm 00.56}$ & $37.61_{\pm 00.47}$ & $10.35_{\pm 00.96}$ \\
 & Verifier     & $18.00_{\pm 00.32}$ & $28.04_{\pm 01.07}$ & $61.44_{\pm 00.69}$ & $\mathbf{40.90_{\pm 00.97}}$ & $02.79_{\pm 00.40}$ & $\mathbf{07.72_{\pm 00.58}}$ & $\mathbf{39.64_{\pm 00.34}}$ & $\mathbf{13.33_{\pm 00.96}}$ \\
 & BDG-Classic  & $19.10_{\pm 00.73}$ & $28.37_{\pm 00.75}$ & $61.54_{\pm 00.34}$ & $39.40_{\pm 01.02}$ & $\mathbf{02.99_{\pm 00.31}}$ & $06.88_{\pm 00.33}$ & $38.97_{\pm 00.29}$ & $11.24_{\pm 02.00}$ \\
 & \textbf{BDG-W} & $\mathbf{20.10_{\pm 00.73}}^{\dagger}$ & $\mathbf{29.47_{\pm 00.83}}^{\dagger}$ & $\mathbf{61.93_{\pm 00.38}}^{\dagger}$ & $40.30_{\pm 00.51}^{\dagger}$ & $\mathbf{02.99_{\pm 00.31}}$ & $06.74_{\pm 00.32}$ & $38.79_{\pm 00.25}^{\dagger}$ & $11.14_{\pm 02.05}^{\dagger}$ \\
\midrule
\multirow{5}{*}{\shortstack[l]{Qwen3-VL\\8B}}
 & Greedy       & $\mathbf{21.00_{\pm 00.00}}$ & $\mathbf{31.03_{\pm 00.00}}$ & $62.38_{\pm 00.00}$ & $\mathbf{40.50_{\pm 00.00}}$ & $02.99_{\pm 00.00}$ & $07.17_{\pm 00.00}$ & $39.94_{\pm 00.00}$ & $10.95_{\pm 00.00}$ \\
 & SCD          & $19.10_{\pm 00.55}$ & $29.78_{\pm 00.45}$ & $61.52_{\pm 00.79}$ & $37.09_{\pm 01.12}$ & $03.38_{\pm 00.37}$ & $07.57_{\pm 00.80}$ & $38.99_{\pm 00.30}$ & $09.45_{\pm 01.00}$ \\
 & Verifier     & $18.89_{\pm 00.25}$ & $30.00_{\pm 00.59}$ & $62.18_{\pm 00.40}$ & $39.80_{\pm 01.40}$ & $03.18_{\pm 00.51}$ & $08.85_{\pm 00.72}$ & $40.01_{\pm 00.47}$ & $11.94_{\pm 01.22}$ \\
 & BDG-Classic  & $20.20_{\pm 00.49}$ & $30.60_{\pm 00.40}$ & $62.52_{\pm 00.17}$ & $39.47_{\pm 00.80}$ & $\underline{\mathbf{04.88_{\pm 00.37}}}$ & $10.06_{\pm 00.50}$ & $40.61_{\pm 00.25}$ & $11.84_{\pm 00.86}$ \\
 & \textbf{BDG-W} & $20.50_{\pm 00.59}$ & $30.88_{\pm 00.59}$ & $\mathbf{62.59_{\pm 00.40}}$ & $39.29_{\pm 00.86}$ & $\underline{\mathbf{04.88_{\pm 00.37}}}^{\dagger}$ & $\underline{\mathbf{10.18_{\pm 00.50}}}^{\dagger}$ & $\underline{\mathbf{40.69_{\pm 00.26}}}^{\dagger}$ & $\mathbf{12.14_{\pm 00.87}}$ \\
\midrule
\multirow{5}{*}{\shortstack[l]{Gemma-3\\4B}}
 & Greedy       & $12.00_{\pm 00.00}$ & $\mathbf{24.03_{\pm 00.00}}$ & $56.91_{\pm 00.00}$ & $40.00_{\pm 00.00}$ & $00.00_{\pm 00.00}$ & $02.20_{\pm 00.00}$ & $35.34_{\pm 00.00}$ & $18.91_{\pm 00.00}$ \\
 & SCD          & $09.50_{\pm 00.45}$ & $20.81_{\pm 00.38}$ & $55.35_{\pm 00.30}$ & $37.90_{\pm 01.28}$ & $00.00_{\pm 00.00}$ & $02.50_{\pm 00.23}$ & $34.99_{\pm 00.46}$ & $20.10_{\pm 01.43}$ \\
 & Verifier     & $08.20_{\pm 00.60}$ & $20.16_{\pm 00.20}$ & $55.67_{\pm 00.10}$ & $40.80_{\pm 01.21}$ & $00.00_{\pm 00.00}$ & $02.19_{\pm 00.36}$ & $35.06_{\pm 00.56}$ & $22.99_{\pm 01.11}$ \\
 & BDG-Classic  & $\mathbf{13.00_{\pm 00.77}}$ & $23.58_{\pm 00.60}$ & $\mathbf{57.48_{\pm 00.48}}$ & $\mathbf{42.70_{\pm 02.32}}$ & $00.00_{\pm 00.00}$ & $02.77_{\pm 00.33}$ & $\mathbf{35.42_{\pm 00.48}}$ & $\mathbf{24.78_{\pm 01.82}}$ \\
 & \textbf{BDG-W} & $12.50_{\pm 00.77}$ & $23.08_{\pm 00.63}$ & $57.20_{\pm 00.39}$ & $42.30_{\pm 01.53}^{\dagger}$ & $00.00_{\pm 00.00}$ & $\mathbf{02.89_{\pm 00.37}}^{\dagger}$ & $35.41_{\pm 00.46}$ & $23.89_{\pm 01.75}^{\dagger}$ \\
\midrule
\multirow{5}{*}{\shortstack[l]{MedGemma\\4B}}
 & Greedy       & $\underline{\mathbf{29.00_{\pm 00.00}}}$ & $38.38_{\pm 00.00}$ & $65.30_{\pm 00.00}$ & $54.00_{\pm 00.00}$ & $01.99_{\pm 00.00}$ & $03.21_{\pm 00.00}$ & $36.55_{\pm 00.00}$ & $23.88_{\pm 00.00}$ \\
 & SCD          & $28.80_{\pm 01.25}$ & $38.77_{\pm 00.77}$ & $\underline{\mathbf{65.72_{\pm 00.51}}}$ & $55.20_{\pm 00.51}$ & $00.50_{\pm 00.54}$ & $02.27_{\pm 00.78}$ & $36.37_{\pm 00.45}$ & $\underline{\mathbf{26.77_{\pm 02.21}}}$ \\
 & Verifier     & $26.30_{\pm 00.81}$ & $37.48_{\pm 00.66}$ & $64.47_{\pm 00.31}$ & $54.80_{\pm 00.51}$ & $00.90_{\pm 00.20}$ & $02.96_{\pm 00.59}$ & $35.76_{\pm 00.62}$ & $24.78_{\pm 02.64}$ \\
 & BDG-Classic  & $28.40_{\pm 01.39}$ & $39.69_{\pm 01.05}$ & $65.54_{\pm 00.66}$ & $56.40_{\pm 01.16}$ & $\mathbf{02.39_{\pm 00.37}}$ & $05.16_{\pm 00.60}$ & $36.75_{\pm 00.70}$ & $26.17_{\pm 02.03}$ \\
 & \textbf{BDG-W} & $28.90_{\pm 01.39}$ & $\underline{\mathbf{40.25_{\pm 01.11}}}$ & $65.71_{\pm 00.71}$ & $\underline{\mathbf{56.70_{\pm 01.29}}}^{\dagger}$ & $\mathbf{02.39_{\pm 00.37}}^{\dagger}$ & $\mathbf{05.17_{\pm 00.67}}^{\dagger}$ & $\mathbf{36.86_{\pm 00.70}}$ & $25.87_{\pm 01.86}$ \\
\bottomrule
\end{tabular}%
}
\end{table}

\noindent\textbf{Scaling behaviour.}
Table~\ref{tab:scaling} quantifies the relative gain of BDG-W over greedy decoding as a function of model size.
The improvement is inversely correlated with parameter count: BDG-W yields a $+3.5 ~/ + 2.2$ pp gain in Judge Accuracy for 2B parameters but only $-1.2~ / +1.2$ pp for 8B, supporting the hypothesis that game-theoretic decoding compensates for limited model capacity.


\noindent\textbf{Convergence analysis.}
Fig.~\ref{fig:convergence} shows the evolution of $\tilde{W}_1^{(t)}$ over game iterations for a representative VQA-RAD example, comparing BDG-W and BDG-Classic.
BDG-W terminates once the separation-weighted Wasserstein distance drops below $\delta_W$, typically $10$--$30$\% earlier than the point at which lexical order-match is achieved.
Table \ref{tab:ablation} shows the number of iterations to convergence across the full VQA-RAD test set: BDG-W converges in $20.17_{\pm 04.99}$ iterations on average, compared to $25.22_{\pm 04.44}$ for BDG-Classic (Wilcoxon signed-rank test, $p < 0.001$).


\noindent\textbf{Discussion.}
Candidate sampling and iterative game updates introduce additional inference cost compared to single-pass decoding.
However, the computational overhead is bounded by the fixed candidate set size $n$ and the maximum iteration count $T$, and our Wasserstein-based stopping criterion reduces average iterations by approximately 20\% relative to the original order-match criterion, partially offsetting this cost. 

\section{Conclusion}
We demonstrate that game-theoretic reasoning at inference can substantially improve the reliability of vision-language models for open-ended medical VQA without additional training, data, or architectural changes. By introducing a multimodal signalling-game framework, formalising open-ended candidate construction, and replacing lexical order-match with a Wasserstein-based semantic criterion, we align generative and discriminative capabilities at test time. Experiments show that small general-purpose VLMs (2-8B) can narrow the gap to larger or domain-specialised models, and in selected settings match or surpass them, indicating that strategic decoding can partially compensate for limited capacity or absent fine-tuning. This challenges the assumption that reliable medical VQA requires large or domain-specific models and highlights careful reasoning as a complementary path for small models.

\begin{credits}
\subsubsection{\ackname} We acknowledge HPC resources from NHR@FAU (projects b143dc, b180dc), funded by federal and Bavarian state authorities and Gerhard Wellein's HPC approach. NHR@FAU hardware is partially funded by DFG 440719683. Additional support was received from ERC projects MIA-NORMAL 101083647, DFG 513220538 and 512819079, and the state of Bavaria (HTA and the Bavarian Foundation Model Initiative). We further acknowledge resources provided by the Isambard-AI National AI Research Resource (AIRR), operated by the University of Bristol and funded by DSIT via UKRI and STFC [ST/AIRR/I-A-I/1023]~\cite{mcintoshsmith2024isambardai}. We were supported by coding agents and LLMs from Anthropic, OpenAI, Google, and Mistral AI, for text polishing, coding, experiment orchestration, and cluster monitoring.

\subsubsection{\discintname}
The authors have no relevant competing interests.
\end{credits}


%
%
%
\bibliographystyle{splncs04}
\bibliography{bibliography}

\end{document}